\documentclass[10pt, conference, compsocconf]{IEEEtran}
\usepackage{cite}
\usepackage{pgf}
\usepackage[cmex10]{amsmath}
\usepackage{amssymb,amsfonts}
\usepackage{algorithmicx}
\usepackage{algcompatible}
\usepackage{algpseudocode}
\usepackage{graphicx}
\usepackage{textcomp}
\usepackage{xcolor}
\usepackage{flushend} % to equalise columns at end
\def\BibTeX{{\rm B\kern-.05em{\sc i\kern-.025em b}\kern-.08em
    T\kern-.1667em\lower.7ex\hbox{E}\kern-.125emX}}
\DeclareUnicodeCharacter{2212}{-} % due to pgf graphs

\begin{document}
\bstctlcite{IEEEexample:BSTcontrol} % shorten authour lists in references

\title{GDDR: GNN-based Data-Driven Routing}

\author{\IEEEauthorblockN{Oliver Hope}
\IEEEauthorblockA{\textit{Computer Laboratory} \\
\textit{University of Cambridge}\\
Cambridge, UK \\
oh260@cantab.ac.uk}
\and
\IEEEauthorblockN{Eiko Yoneki}
\IEEEauthorblockA{\textit{Computer Laboratory} \\
\textit{University of Cambridge}\\
Cambridge, UK \\
eiko.yoneki@cl.cam.ac.uk}
}

\maketitle

\begin{abstract}
We explore the feasibility of combining Graph Neural Network-based policy architectures with Deep Reinforcement Learning as an approach to problems in systems. This fits particularly well with operations on networks, which naturally take the form of graphs. As a case study, we take the idea of data-driven routing in intradomain traffic engineering, whereby the routing of data in a network can be managed taking into account the data itself. The particular subproblem which we examine is minimising link congestion in networks using knowledge of historic traffic flows. We show through experiments that an approach using Graph Neural Networks (GNNs) performs at least as well as previous work using Multilayer Perceptron architectures. GNNs have the added benefit that they allow for the generalisation of trained agents to different network topologies with no extra work. Furthermore, we believe that this technique is applicable to a far wider selection of problems in systems research.
\end{abstract}


\section{Introduction}
Recently, Deep Reinforcement Learning (DRL)\cite{sutton2018reinforcement} techniques have gained traction in the field of systems research as regarding performing optimisations to allow them to run faster or more efficiently. Traditionally, this kind of work has been performed by exploiting expert knowledge, random search, or Bayesian Optimisation (BO). Recently, the power of machine learning (ML) approaches has been demonstrated by auto-tuners using BO. Many modern problems in computer systems feature both dynamic inputs and are combinatorial in nature, suggesting DRL as a potential approach.

Many systems can be modelled as graphs (most obviously in the case of computer networks). In these cases, the systems being operated on by DRL can change size and structure dramatically without actually impacting the logic or the structure of the problem being solved itself. However, many designs of agents in DRL rely on fixed input and output sizes, which can cause issues. These issues can only be solved with clever policy design. Such `clever' design often implies incorrect relationships and biases in the structure of the problem\cite{47094}. A recent development in ML is that of Graph Neural Networks (GNNs)\cite{4700287}. We believe these can be successfully combined with DRL techniques to allow for generalisation of agents to different graphs without inducing extra overheads and avoiding laborious tinkering with policy design.

To explore the issue presented above and the effectiveness of GNNs as a solution, we have taken the case study of data-driven routing in intradomain traffic engineering (TE). Routing is an integral part of the functioning of almost all modern computer networks. In particular, the internet consists of many different Autonomous Systems (ASs) which must both route data internally and between each other. In this work we focus on the problem of routing data inside these systems. When data is to be sent from one node to another we must first choose a path to send it on, as there are generally multiple different paths that a single packet could take across the network. One potential path selection method would aim to reduce link congestion across the network. But for it to perform optimally we must have complete knowledge of the traffic demands across the network in advance, which is rarely the case.

The problem of knowing the traffic demands across a network in advance to perform good routing can be solved in part with good enough predictions. This has been done using both Supervised Learning and, with more success, DRL techniques by Valadarsky et al.\cite{valadarsky2017learning}, who used a multilayer perceptron (MLP)\cite{HORNIK1989359} for their DRL architecture. The benefits given by using DRL here are that it allows routing strategies to be learnt in a way not possible with supervised learning as we can avoid explicitly predicting demands as a substep. Additionally, scaling to real-world systems and distributed systems should be more straight-forward.

However, this solution was not generalisable, meaning that the agent would need to be retrained from scratch for every different network it is used on. This is a greater problem when one takes into account that networks often undergo small changes, as they are reconfigured often. This is why in this work we propose to extend the DRL technique using GNNs (creating GDDR: GNN-based Data-Driven Routing) which will allow us to generalise the learning across different network topologies.

We begin by providing a detailed description of the case study including its context, design, and results before providing a wider discussion of what this means for GNNs and their generalisability when used with DRL techniques on networks.

\subsection{Contributions}

In summary, this paper makes the following research contributions:
\begin{itemize}
  \item We show that GNNs can be combined with DRL techniques and applied to systems such that generalisation is possible. This means that agents do not need to be retrained across different environments in the same general domain.
  \item We show that GNNs do not cause extra overhead (in terms of learning time and performance) compared with MLPs when used for routing. (See section \ref{section:evaluation})
  \item We show that one can use GNNs with DRL on networks to decrease the effort required when designing policies that must be generalisable. (See section \ref{section:discussion})
  \item Using intradomain TE as a case study, we introduce different policy designs and techniques for approaching data-driven routing in a way that is generalisable to different network topologies and traffic types. (See section \ref{section:policies})
  \item We provide an environment to be used for further experimentation with DRL in data-driven routing. (See section \ref{section:environment})
\end{itemize}

\section{Intradomain Traffic Engineering and the Routing Problem}

Traffic engineering, as previously mentioned, contains the problem of how to control and manage computer networks. With intradomain routing, we focus on which paths data should take on a single network controlled by a single entity. There are many benefits of working in this space. We can make many assumptions such as correct knowledge about the topology of the network and the bandwidth of its links. We can also assume to have some level of singular control over the routing decisions made within it. Neither of these assumptions can be made in interdomain routing. However, this is still a relevant problem as the internet does consist of many interconnected ASs which must all route internally.

When deciding on routing policy inside a network there are many possible choices, and a method providing optimality is preferred. We can of course choose to optimise different things. For instance, reducing latency, in which case a sensible choice would be shortest-path routing policies. Examples of widely-used policies for this purpose are OSPF\cite{rfc5340} and RIP\cite{rfc4080}. However, often the best utility for many clients on a network can be achieved by reducing congestion. Here, we aim to reduce the number of flows that experience packet loss and so are unnecessarily slowed. One way to approach such a problem is by aiming to minimise the maximum congestion on any link across a network.

\subsection{Multicommodity Flow}
\label{section:multicommodity}

Many policies plan routes based on the physical state of the network, taking into account details such as the network topology and the bandwidth of its links\cite{Barnhart2001}. However, this is not enough information to provide close-to-optimal congestion avoidance with routing strategies. It is already known that if we model the network as a flow network and have complete knowledge of traffic requirements in advance, then we can solve the multicommodity flow problem to calculate optimal routes\cite{10.5555/580470}.

The problem is set up as follows: The network is modelled as a flow network $G(V, E)$ where each edge $(u, v) \in E$ has a capacity $c(u, v)$. Each flow is then a `commodity', $K_i = (s_i, t_i, d_i)$ on the network where $s_i$ is the source, $t_i$ the sink, and $d_i$ the demand (the quantity of data in that flow). $f_i(u, v)$ denotes the fraction of flow $i$ along edge $(u, v)$ and $k$ is the number of flows. With these definitions we can then set up several constraints that must hold on the network:

\begin{enumerate}
    \item Flow on a link must not exceed its capacity:\\
    $\forall (u,v)\in E:\,\sum_{i=1}^{k} f_i(u,v)\cdot d_i \leq c(u,v)$
    \item Flow entering and exiting a node must be conserved:\\
    $\sum_{w \in V} f_i(u,w) - \sum_{w \in V} f_i(w,u) = 0, \quad u \neq s_i, t_i$
    \item The entire flow of a commodity must exit its source:\\
    $\sum_{w \in V} f_i(s_i,w) - \sum_{w \in V} f_i(w,s_i) = 1$
    \item The entire flow of a commodity must be absorbed at the sink:\\
    $\sum_{w \in V} f_i(w,t_i) - \sum_{w \in V} f_i(t_i,w) = 1$
\end{enumerate}

Within this framework we can specify any valid routing we wish: a flow/commodity is able to take any path and be split across many paths. In this system it is possible to calculate a routing that optimally minimises congestion when one knows data about commodities such as the demands in advance. However, this begs the all important question: where do we get the data on future traffic flows? The unfortunate answer in most cases is that it simply does not exist. One possible solution is to predict future demands and then derive routings by solving the multicommodity flow problem. However, this does not lead to good results when the predictions are incorrect. Generally, shortest-path routing is more successful. Another solution is to use `Oblivious routing' schemes\cite{bansal2008} (discussed in section \ref{section:oblivious}). These aim to obtain good performance under any kind of traffic demand and they perform well, but not optimally.

\section{Routing with Reinforcement Learning}
\label{section:rl}

Good routing strategies can be found if we know the exact details of future traffic data. However, predictions of such data lead to bad results even if they are only partially wrong. Therefore, an alternative solution is to try instead to directly predict a routing strategy without a traffic prediction substep. This is the method selected by Valadarsky et al. in their paper `Learning To Route with Deep RL'\cite{valadarsky2017learning}, the inspiration for much of the work presented here.

In that paper, the authors realised that the traffic patterns in many networks tend to have temporal regularities, meaning that similar patterns tend to reoccur. This is of course unsurprising as networks are used by people who tend to live by cyclic patterns (weeks, days, etc.). This means we can use historical traffic data as a predictor of future traffic data, as long as our assumption that regularities exist in the traffic holds. The authors then used a DRL technique to predict good future routing strategies from historical demand data.

Reinforcement Learning\cite{sutton2018reinforcement} is one branch of Machine Learning which models agents taking actions in a pre-specified environment, similar to how people take actions in the world. More formally, we have the 4-tuple $(S, A, P_a, R_a)$ where:
\begin{itemize}
    \item $S$ is a finite set of states
    \item $A$ is a finite set of actions
    \item $P_a(s,s')$ is the probability that action $a$ in state $s$ at time $t$ will lead to state $s'$ at time $t + 1$
    \item $R_a$ is the reward received after transitioning from state $s$ to state $s'$, due to taking action $a$
\end{itemize}
The agent receives states from the environment (and rewards to tell it how good its choices have been) and returns actions to be taken in that environment. This process then repeats until some stopping condition is met.

Valadarsky et al.\cite{valadarsky2017learning} model the network as an environment where each environment step is a discrete timestep. The observation in each step is a history of traffic demands in the network given as matrices. Their agent learns how to provide a good routing strategy as an action. This method achieved results that approached the optimal routing which is only possible if one were to have had perfect future knowledge.

They created a routing strategy as an output action from their agent using a method they introduced called `softmin routing'. We took this method, extending and improving it, which is explained in section \ref{section:softmin}.

Their agent was a simple multilayer perceptron (MLP) design which provided good performance with little need for tuning. However, their method is not generalisable because the input and output sizes of an MLP are fixed. Both the observation and action sizes depend on the size of the network, therefore it must be retrained with a new agent for every different network topology.

\section{Routing with Graph Neural Networks}
\label{section:graph_neural_networks}

The problem of generalisation as mentioned previously hinges on the fixed-size nature of the MLP. To avoid this issue we instead built our DRL agent around a Graph Neural Network (GNN)\cite{4700287}.
GNNs are a form of neural network that has gained considerable traction over the last few years. This is because, in the same way that convolutional neural networks (CNNs) have been used with much success on images (pixels form a grid and are often related to their neighbouring pixels), many other datasets are graph-structured with relations to neighbouring nodes being an essential part of their structure. GNNs effectively allow us to generalise the CNN model onto graphs. There are many different types of GNN with different trade-offs, but for our research we have used the most general model to date: that proposed by Battaglia et al.\cite{47094}.

In this model, a graph network (GN) block is the central item which has graphs as inputs and outputs. Here a graph is defined to be the 3-tuple $G = (u, V, E)$ where $u$ is a global attribute vector, $V = \{v_i\}_{i=1:N^v}$ is the set of vertex attribute vectors, and $E = \{(e_k, r_k, s_k)\}_{k=1:N^e}$ is the set of edge attribute vectors and their source and destination vertices. The GN block itself contains three $\phi$ functions which update attribute information and three $\rho$ functions which pool attribute information to be used in these updates. It is the parameters of these six functions that can be learnt, and the flexibility to modify these functions that makes this form of GNN so general. In this work we implement all of these functions as MLPs.

\subsection{Formal Routing Specification}
\label{section:routing}

Now that we have described all the necessary building blocks (a problem: minimising congestion in intradomain routing, a method: DRL with GNNs) we need a formal specification to work to and around which to build our solution. For this we again took our inspiration from Valadarsky et al. and the multicommodity flow problem.

We consider a static network upon which we can specify a routing strategy, and which also has a set of traffic demands associated with it in every timestep in the form of matrices. We define:

\begin{itemize}
  \item The \emph{network}, which is modelled as a directed graph where all the edges have a link capacity: $G=(V,E,c)$ where $V$ is the set of vertices, $E$ is the set of edges and $c : E \rightarrow \mathbb{R}^+$ is a function mapping each edge in the graph to its capacity.
  \item The \emph{routing}, which for each flow (demand, source, and destination pair) specifies at each vertex how much of that flow should be sent down each of its edges to each of its neighbours. Therefore, if we define $\Gamma(v)$ to be the set of all neighbours of vertex $v$ then we can define a routing to be $\mathcal{R}_{v,(s,t)} : \Gamma(v) \rightarrow [0,1]$ with $\mathcal{R}_{v,(s,t)}(u)$ as the proportion of the flow passing from $s$ to $t$ through vertex $v$ that is forwarded to vertex $u$. Importantly, any routing specified must obey the two constraints:
    \begin{enumerate}
      \item No traffic is lost between source and destination:\\
        $\sum_{u \in \Gamma(v)}{\mathcal{R}_{v,(s,t)}(u)} = 1 \qquad \forall s, t \in V \wedge v \neq t$
      \item All traffic for a destination is absorbed at that destination:\\
        $\sum_{u \in \Gamma(v)}{\mathcal{R}_{t,(s,t)}(u)} = 0 \qquad \forall s, t \in V$
    \end{enumerate}
  \item The \emph{demands} which can be represented as matrix $D \in \mathbb{R}^{|V|\times|V|}$ called a demand matrix (DM) where each element $D_{st}$ is the traffic demand between the source $s$ and destination $t$.
\end{itemize}

With these definitions we can now describe the movement of traffic over a network in enough detail to examine how to perform routing strategies with DRL. The remainder of this work uses this system for all experiments and models. In examples, we will assume that the network itself is fixed and that traffic across it is described by sequences of DMs, also fixed, each of which represents a discrete timestep. The only part that we (or the agent) is allowed to modify is the routing strategy. As the aim is to provide a better routing strategy we need to translate our notion of optimality into this system's equations. As previously described, we aim to reduce congestion by minimising the link over-utilisation on the network. To allow solving for this to be feasible, we linearise the utility function. This can be defined as minimising $U_{max}$ in Equation~\ref{equation:utility} where $U(u, v)$ is the utilisation of the link $(u, v)$.
\begin{equation}
  \label{equation:utility}
  \forall (u,v) \in E: U_{max} > U(u,v)
\end{equation}
Therefore, we can calculate a $U_{max}$ for each DM over a routing strategy and the agent can use this to learn.

\section{Environment Design}
\label{section:environment}

\begin{figure*}
  \centering
  \includegraphics[width=0.7\textwidth]{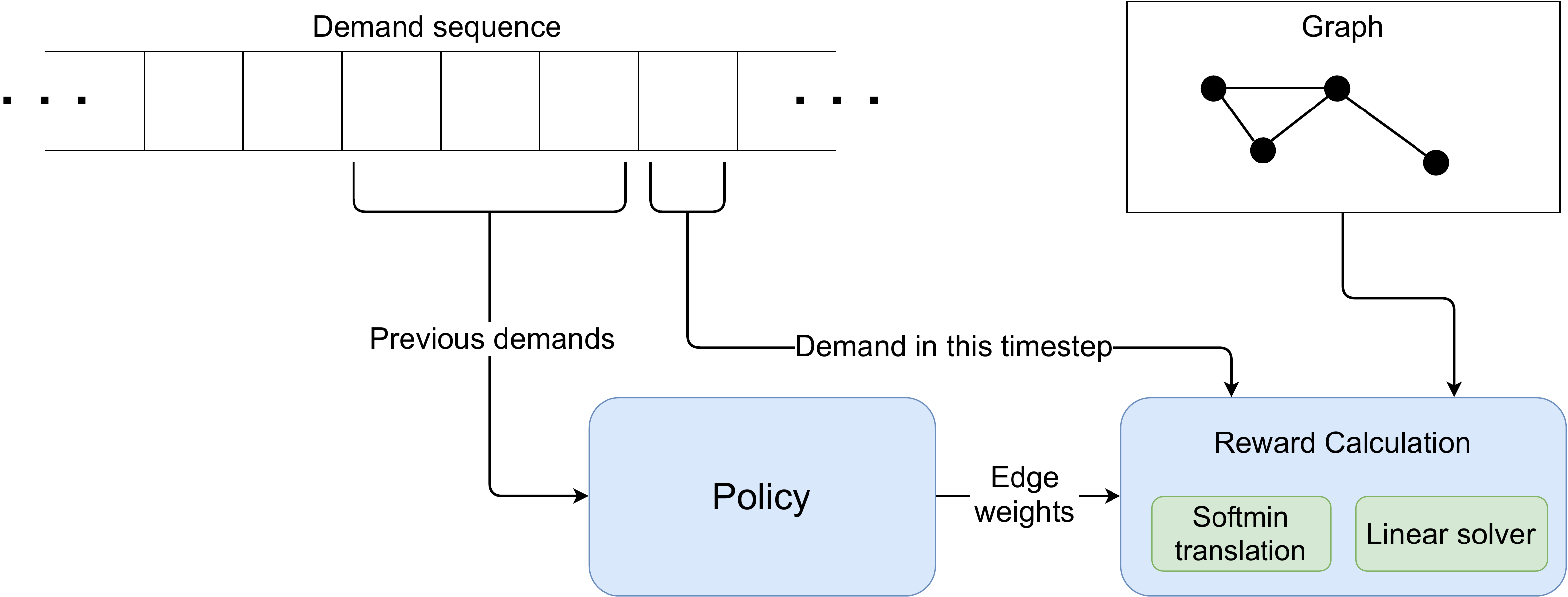}
  \caption{Environment dataflow in a single timestep. Top left is the generated demand sequence, and top right is the graph we are routing over. We can see that the previous $n$ demands (here 3) are given to the policy as input while the reward is calculated using the new demand for this timestep.}
  \label{fig:environment}
\end{figure*}

So that it is possible to both train and evaluate approaches to the problem, it was necessary first to build an RL environment to simulate the desired responses with which an RL agent could interact. For easy interoperability with existing libraries, we decided that this environment should have an OpenAI Gym\cite{1606.01540} API. A diagram of the overall structure of the environment is given in Figure~\ref{fig:environment}. We will now proceed to a discussion on how we designed the interactions.

\subsection{Reward Calculation}

As the optimal routing that can be achieved on a given graph varies for different demand matrices, the reward cannot be derived solely from the calculation of the maximum link utilisation. Fortunately, as described in section~\ref{section:multicommodity}, we know that an optimal routing does exist and it can be found in polynomial time with linear programming (LP). To do this, we solve the problem using the standard LP formulation, minimising the utility function given in Equation~\ref{equation:utility}. The environment implements a linear solver for the optimal routing to calculate the optimal link utilisation. We implement this solver on top of Google OR-Tools\cite{ortools}. Then, the reward is derived by comparing the routing strategy produced by the DRL agent to the calculated optimal routing. The original work used Equation~\ref{equation:reward}, where $U_{max}$ is the maximum link utilisation and we use the same.
\begin{equation}
  \label{equation:reward}
  \mathrm{reward} = -\frac{U_{max_{agent}}}{U_{max_{optimal}}}
\end{equation}

\subsection{Observation space}

In Valadarsky et al.'s work\cite{valadarsky2017learning}, an observation is a history of traffic demands, which is presented as a list of traffic matrices and then flattened for input to the multilayer perceptron (MLP). However, our work aims to make the solution generalisable over graphs of different shapes using GNNs. As GNNs work on graphs, it is possible to vary both the number of vertices and edges that are input to a GNN (something not possible with an MLP). Unfortunately, this previous method of managing demands as observations no longer works. The reason for this is that one way to input the demands to the GNN would be to place the demands associated with each vertex on that vertex as vertex attributes. The issue here being that the number of demands associated with a vertex scales with the number of nodes in the graph. Therefore, the size of node attributes in the GNN would have to grow as more vertices are added, which is unfortunately not possible within the structure (we are able to input to more nodes but the size of individual node features must remain constant).

This issue required us to find a way to associate the demands with the correct nodes such that the agent is still able to learn but only requires a constant amount of space per-vertex as the graph grows. The solution we decided upon was summing the total outgoing flow and incoming flow for each vertex, meaning that the observation size becomes $O(|V|)$ as opposed to $O(|V|^2)$ and so can be used with a GNN.

One other important addition enabling this new structure to work was normalising the inputs, as otherwise the more vertices in a graph, the greater the size of the input features, which could lead to unwanted behaviour.

\subsection{Action Space}

Following on from the definitions given in section~\ref{section:routing}, we can see that one valid way for the agent to assign a routing would be to provide splitting ratios for each edge under each flow. However, this would require an output of $|V|\times(|V|-1)\times|E|$ separate values. Unfortunately, this size of action space is too big to enable successful learning. If we make several approximating assumptions about the routing (which will no longer allow us to achieve the optimal routing but still allow us to get closer than oblivious strategies), then we can reduce this output size.

A first method to reduce the action space size is to ignore the source of any packets, forming a destination-only routing. This method reduces the size to $|V|\times|E|$. However, this is still too large, so we decided instead to create a method of deriving routing strategies from setting weights of edges, creating an action space with only $|E|$ values. This space was finally small enough to achieve good results. The next section describes how the routing strategy works.

\section{Routing Translation}
\label{section:softmin}

Valadarsky et al.\cite{valadarsky2017learning} confronted the same issues of action space size as we did. That is, the issue where the number of different possible routing decisions to be made, if one allows routing weights to be specified for all flows, is too great and so leads to the agent struggling to learn. Therefore, they created a method of deriving a routing strategy from edge weights to make it smaller, which they called \emph{softmin routing}. As part of our research, we attempted to use the same solution specified but found several issues relating to routing loops and therefore made modifications. It is this modified softmin routing that we will present below.

Softmin routing is a way of deriving the splitting ratios on each edge from each vertex, per-flow, given weights that have been set on each edge. These values are calculated using algorithm~\ref{algorithm:softmin} which we shall now describe. We calculate the ratios per-flow, where a flow is a source-destination pair, $(s,t)$. For each vertex, we calculate its distance to the destination vertex (along its shortest path using the weighted edges). Then, for each vertex, we add the weight of each outgoing edge to the distance of the neighbour at the end of that edge. Using the softmin function given in Equation~\ref{equation:softmin} with these summed numbers as input, we are returned splitting ratios to use on these edges for routing under this particular flow. We then repeat this process for all vertices and all flows. The softmin function takes a vector and normalises the values to a probability distribution. It also has a parameter, $\gamma$, which varies how spread the probabilities are in the result.

\begin{equation}
  \label{equation:softmin}
  \operatorname{softmin}(\boldsymbol{x}) = \left(\frac{e^{-\gamma \boldsymbol{x}_i}}{\sum_{j}{e^{-\gamma \boldsymbol{x}_j}}}\right)_i
\end{equation}

One will notice from this description that the routing derived from such a scheme does contain a potential flaw. Although it follows all the rules specified for a routing strategy in section~\ref{section:routing} (no traffic is lost and sources send the required demand which is all absorbed by the destination), there is nothing to stop routing loops occurring. This situation is undesirable for two reasons. First, it wastes capacity as it means traffic traces the same route more than once; second, it increases latency (this is generally unacceptable but is not under measurement here so does not impact our goals, although it would be problematic in a real-world scenario). Therefore, to achieve good results, we have to break routing loops. We do this by converting the graph to a directed acyclic graph (DAG).

\begin{figure}[t]
\small
\algrenewcommand\algorithmicindent{0.5em}
\begin{algorithmic}
\Function{SoftminRouting}{$G$, $\boldsymbol{w}$, $\gamma$}
  \For{$(s, t) \in flows$}
  \State\Comment{Convert to a DAG for the source-sink pair}
    \State $G \gets \operatorname{PruneGraph}(G, (s, t), \boldsymbol{w})$
    \For{$v \in V$}
      \State\Comment{Find the distance of each vertex to the sink}
      \State $d[i] \gets \operatorname{ShortestPath}(i, t)$
    \EndFor
    \For{$v \in V$}
      \State $out\_edges \gets \operatorname{GetOutEdges}(v)$
      \State $out\_weights \gets \{w[(u, v)] + d[v] | (u, v) \in out\_edges\}$
      \State\Comment{Edge length $+$ neighbour's distance}
      \State $softmin\_weights \gets \operatorname{Softmin}(out\_weights, \gamma)$
      \For{$e \in out\_edges$}
        \State $splitting\_ratios[e] \gets softmin\_weights[e]$
      \EndFor
    \EndFor
  \EndFor
  \State \textbf{return} $splitting\_ratios$
\EndFunction
\end{algorithmic}
\caption{Softmin routing algorithm: steps taken to convert learned edge weights given by an DRL agent into a fully-defined routing strategy.}
\label{algorithm:softmin}
\end{figure}

The breaking of loops is a problem often faced in routing. In many protocols such as shortest-path based protocols, it is generally not an issue, as the shortest path by definition cannot contain any cycles. This is not always true as distance vector protocols can produce issues when the network state has not yet converged. It would be easy to remove all loops between source and destination by only keeping the shortest path. However, this does not assist with load-balancing and so is not a useful strategy here. We aim to take advantage of multipath with our routings, so the longer paths need to remain in the network.

The strategy we devised to remove loops is shown in algorithm~\ref{algorithm:pruning} which we will now describe. The input to the algorithm is the graph with edges weighted by the agent. We begin by running Dijkstra's algorithm from the source, recording for each node its parent (in the case of the sink, multiple parents) and any locations where the frontier hits an already explored vertex, called \emph{frontier meets}. Then we trace back from the sink to source, following parent links, and marking any vertex we pass through as \emph{on path}. At this point, we have only recorded any shortest paths. Therefore, for every \emph{frontier meet}, we find the distance to sink of the first ancestor from each side of the edge to the sink. We then update all the parent information and \emph{on path} information for these vertices so that it becomes a valid path from the more distant ancestor to the closer ancestor. Finally, we remove all edges between nodes that are not \emph{on path} and all edges that are \emph{on path} but where the head of the edge is at the parent.

\begin{figure}[t]
\small
\algrenewcommand\algorithmicindent{0.5em}
\begin{algorithmic}
\Function{PruneGraph}{$(V, E, W), (s,t)$}
  \State $queue \gets [(0, s, [])]$
  \Comment{Queue element is: (distance, vertex, parents)}
  \While{$|queue| \not= 0$}
    \Comment{Run Dijkstra's algorithm}
    \State $(d, v, \boldsymbol{p}) \gets \operatorname{pop}(queue)$
    \State $parents[v] \gets \boldsymbol{p}$
    \For{$u \in \operatorname{neighbours}(E, v) - \boldsymbol{p}$}
      \If{$u = t$}
        \Comment{Stop when reach sink}
        \State $\operatorname{append}(parents[t], v)$
      \ElsIf{$u \in explored$}
        \State\Comment{Record when see already explored node}
        \State $\operatorname{append}(frontier\_meets, (v, u))$
      \Else
        \State $\operatorname{push}(queue, (d + W(u,v), u, [v]))$
      \EndIf
    \EndFor
    \State $\operatorname{append}(explored, v)$
  \EndWhile
  \State

  \State $queue \gets [t]$
  \While{$|queue| \not= 0$}
  \State\Comment{BFS to mark vertices on path source to sink}
    \State $v \gets \operatorname{pop}(queue)$
    \State $\operatorname{append}(on\_path, v)$
    \For{$p \in parents[v]$}
      \State $d[p] \gets d[v] + G[p][v]$
    \EndFor
  \EndWhile
  \State

  \For{$(u,v) \in frontier\_meets$}
    \State\Comment{Assign paths at frontier collisions}
    \State $a \gets \operatorname{ancestor}(E, on\_path, u)$
    \State $b \gets \operatorname{ancestor}(E, on\_path, v)$
    \If{$d[a] = d[b]$}
      \State \textbf{continue}
    \EndIf
    \State $\operatorname{ConnectPath}(a, u, v, b)$
    \Comment{Update parent pointers for new path}
  \EndFor
  \State

  \For{$(u,v) \in E$}
    \Comment{Remove edges not leading to sink}
    \If{$u \notin on\_path \wedge v \notin on\_path \wedge u \notin parents[v]$}
      \State $E \gets E - \{(u, v)\}$
    \EndIf
  \EndFor
  \State \textbf{return} $(V, E)$
\EndFunction
\end{algorithmic}
\caption{DAG conversion algorithm retaining high path count from source to sink}
\label{algorithm:pruning}
\end{figure}

\section{Routing Agent Policy Design}
\label{section:policies}

Reinforcement Learning, as introduced in section \ref{section:rl}, requires an agent, an environment, and a specification for how the two interact. Thus far we have presented all of these apart from the design of the agent itself. While the environment must be a good mapping from reality to make sure one is actually solving the problem one claims to be, its design consists of a series of simplifying assumptions. The interface design is important because a bad interface can lead to learning being slow or impossible. However, the most important puzzle piece is the policy chosen for the agent to use when making decisions. The simplest agent that could be chosen for a DRL system would be an MLP, which is exactly what was chosen in previous work and proved effective.

\begin{figure}
    \centering
    \includegraphics[width=0.35\textwidth]{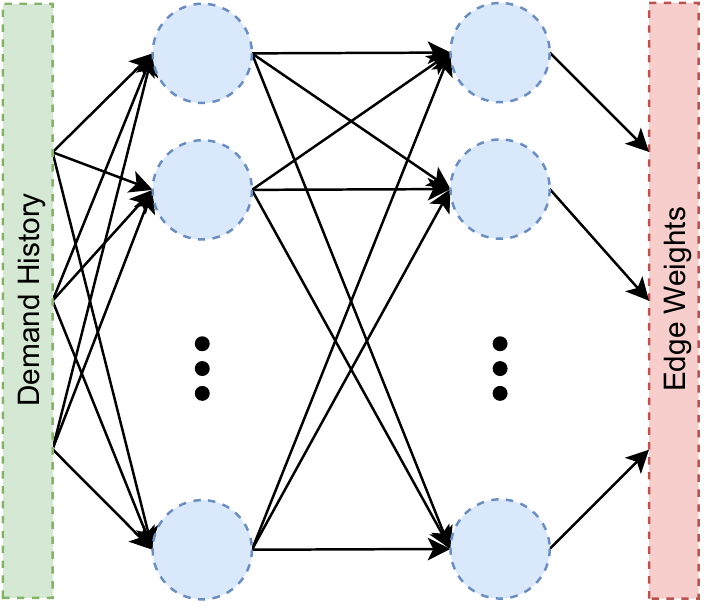}
    \caption{The structure of the MLP policies for data-driven routing.}
    \label{fig:mlp}
\end{figure}

An example of an MLP is given in Figure~\ref{fig:mlp}, where we can see multiple fully connected layers with an input layer on the left and an output layer on the right. The issue for our problem is that we cannot easily change the size of the input (or output). We can reduce the input size by ignoring inputs, but this is wasteful. More problematically, we cannot make the input size larger. In the next section we will show that the GNN architecture does not suffer from the same issue.

\subsection{GNN Policy}
\label{section:gnn_policy}
The new work is introduced here with the application of GNNs to the problem. The first stage was to make sure that a GNN can perform routing to the same level as the MLP. To achieve this, we began with the constrained space where we read out an entire routing from the GNN policy as a single action. Due to the nature of training frameworks, this only allows us to train the network on one graph, as the output size must be fixed and is the number of edges of the graph. However, it does still allow us to apply the trained model to different graphs.

There are many different types of GNN to choose from. For example, there is the Graph Attention Network\cite{velickovic2018graph} which uses self-attention when propagating and is parallelisable, and the Message Passing Neural Network\cite{10.5555/3305381.3305512}, which first generalised different types of GNN that pass data along edges and combine it in some way. In this specific problem, we need outputs on edges and inputs on nodes, so the model must derive edge features taking into account node features. It is also the case that the volume of demand to and from a specific node can have a global impact, as such a flow may be destined for the most distant vertex from this one. Therefore, the structure will have to be able to take into account global information when setting edge weights. Finally, the structure of the graph in the way its edges connect has a significant impact on how flows can be routed and their resulting effect on utilisation. To take advantage of this, our policy must be able to see how information flows and so must make use of the edges. Given these requirements, it makes sense to use one of the most general kinds of GNN so as not to place accidental restrictions on learning.

\begin{figure*}
    \centering
    \includegraphics[width=0.85\textwidth]{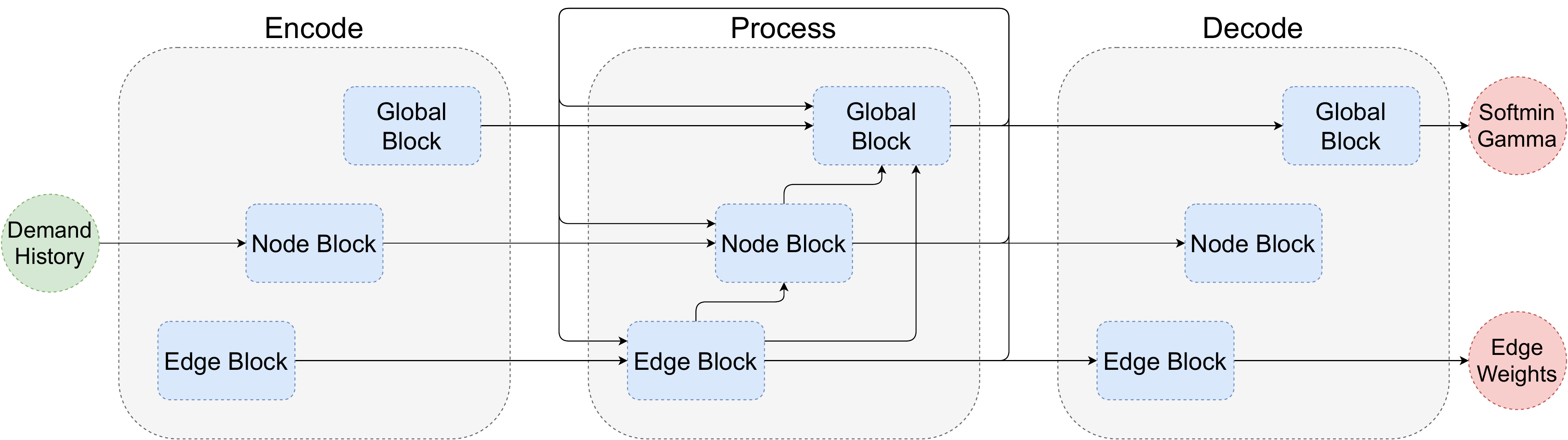}
    \caption{The encode-process-decode GNN structure. On the left are the inputs to the policy, followed by the attribute encoding block. Then we have the full graph network block which performs the message passing and the core of the computation. Finally on the right is the attribute decoding block and the policy outputs. We can see an extra loop from output to input on the central block, which is used in the case that we have multiple message-passing steps to carry back the updated state.}
    \label{fig:encode_process_decode}
\end{figure*}

In keeping with the above observations, we settled on using a fully connected graph network block as defined by Battaglia et al.\cite{47094} and shown in section~\ref{section:graph_neural_networks}, as this is the most general method achievable with the framework and means that we have not constricted any information flows unnecessarily. We are also aware that constraining the feature vectors for edges and vertices to length one and two respectively, due to the input and output sizes, may make learning less efficient and results further from the optimal. Therefore, we expand from the core processing GN block to an `encode-process-decode' model\cite{47094}. This model consists of first mapping all the node features using a learned function to larger hidden size, then performing several computation steps by the core network, and finally mapping all of the edge features to the correct output size using another learned function. The structure of this policy can be seen in Figure~\ref{fig:encode_process_decode}.

Inside the graph network blocks, we use MLPs with learned parameters for all the attribute update $\phi$ functions (separate MLPs for updating the edge, vertex, and global attributes). For the pooling $\rho$ functions, we use the \texttt{tf.unsorted\_segment\_sum} function from TensorFlow to aggregate attributes across vertices and edges.

In summary, the inputs are incoming and outgoing demands for each vertex:
\begin{equation}
  \label{equation:node_inputs}
  \boldsymbol{V} = \left\{\left(\sum_{j=1}^{|V|}{D_{ij}}, \sum_{j=1}^{|V|}{D_{ji}}\right)\right\}_{i=1:|V|}
\end{equation}

and the outputs are a weight for each edge:
\begin{equation}
  \label{equation:edge_outputs}
  \boldsymbol{E} = \{w\}_{i=1:|E|}
\end{equation}

\subsection{Iterative GNN Policy}
\label{section:gnn_iterative}
Given the caveats of the basic GNN policy described above, we felt it necessary to design a policy that can be trained and run on different graphs. This leads to better generalisation. For this to work, we need a fixed-size action space. Fortunately, similar problems have been encountered before, so we were able to take inspiration from works such as Placeto\cite{addanki2019placeto} to create an iterative DRL approach to routing.

In the iterative approach, we only output the value for a single edge at each action, meaning that the network has to be able to output the value for the right edge in the correct action step. The way this is implemented is by providing extra information in the observation to help with the embedding. In Placeto\cite{addanki2019placeto}, Ravichandra et al.\ use a Recurrent Neural Network (RNN) with a hand-designed method of pooling and message-passing steps to build a graph embedding feature vector that can be passed as input to the RNN in each step. We instead harness the more recent research on GNNs, letting a GNN learn the embedding itself. To do this, we modified the simple GNN policy introduced above.

For this policy, getting a set of edge weights for a routing strategy given a demand history takes place in the course of a set of iterations. This is because we can only set one edge value at a time and there are multiple edges. To enable this, we encode the information of which edge we wish to set on the inputs. As before, the node attribute inputs are the demand history, as shown in Equation~\ref{equation:node_inputs}---however, this time we also create input attributes for the edges. The input attribute vector per edge becomes the 3-tuple defined in Equation~\ref{equation:edge_inputs}. That is, a value in the interval $[-1,1]$ denoting the current value for this edge in the routing (and 0 if not set), a binary value denoting whether this edge's value has already been set in the routing, and another binary value denoting whether this is the edge to be set in the current iteration or not.

\begin{equation}
  \label{equation:edge_inputs}
  E = \left\{ (\operatorname{weight}_i, \operatorname{set\vphantom{_i}}_i, \operatorname{target}_i) \right\}_{i=1:|E|}
\end{equation}

These observation inputs are passed into the same encode-process-decode model described in section~\ref{section:gnn_policy}. As the output is to be only for one edge, we read it not from the output edge attributes but the output global attributes. In this policy, the global attribute output is the 2-tuple defined in Equation~\ref{equation:global_outputs} where the first element is the value to set on the edge specified and the second is the value of $\gamma$ to use for the softmin routing (although this is only read on the last iteration for a particular demand history).

\begin{equation}
  \label{equation:global_outputs}
  U = (\operatorname{weight}, \gamma)
\end{equation}

\section{Evaluation}
\label{section:evaluation}
The work thus far has motivated the use of GNNs to allow for generalisation when used with DRL on networks. In particular we have delved into the problem of intradomain TE as an example. We now present the results of experiments to show the benefits that the GNN approach brings. These experiments aim to show:
\begin{itemize}
  \item That GNN-based methods have no performance overhead compared to the MLP method.
  \item That GNN-based methods have no learning overhead compared to the MLP method.
  \item That GNN-based methods are able to generalise to different topologies in routing.
\end{itemize}

\subsection{Baselines}
To quantify the success of our experiments, using a baseline for comparison was necessary. To do this we implemented the simple MLP agent specified by Valadarsky et al.\cite{valadarsky2017learning}. As described above, our policies aim to perform at least as well as these ones for fixed networks. For the generalisation case where we use multiple topologies, the MLP agent cannot be applied so instead we show that generalisation does occur and compare our two different GNN-based policies.

We have also used shortest-path routing as another comparison for all of the tests to show how well the DRL agents perform in comparison to a simple classical method.

\subsection{Traffic and Network Data}
\label{section:traffic_generation}

The topologies we used for testing and training were taken from `The Internet Topology Zoo'\cite{6027859} so that they are representative of the real world. Graphs from this set were used for both testing and evaluation (sometimes with modifications).

The way that traffic demand sequences were made for training and testing was more artificial as our assumption was that there would be some form of temporal regularity in the demand sequences. It was easiest to force this by generating sequences ourselves. To do this we used techniques suggested by Valadarsky et al.\cite{valadarsky2017learning} and inspired by temporal consistencies discussed in \cite{1003042}. These were to create sequences of repeating cycles of demands (cyclical sequences). The demands themselves were generated by being from one of two distributions to simulate occasional `elephant flows' on the network (bimodal demands). The formal definitions for these sequences are:
\begin{itemize}
  \item Bimodal DM (parameters given by example values):\\
    $D_{ij} = p$ if $s > 0.8$ else $q$ where $p \sim \mathcal{N}(400, 100), q \sim \mathcal{N}(800, 100), s \sim \mathcal{U}(0,1)$.
  \item Cyclical sequence:\\
    $\boldsymbol{x} = \left\{ D_{i \mod q} \right\}_{i}$ where $D$ is a sequence of $q$ DMs.
\end{itemize}

\subsection{Training}

Training was performed using the environment which was implemented in Python, using the NetworkX library\cite{SciPyProceedings_11} for graph operations and TensorFlow\cite{tensorflow2015-whitepaper} as the machine learning framework. For the DRL algorithm we decided to use Proximal Policy Optimisation (PPO)\cite{schulman2017proximal} in the form of the PPO2 implementation from the stable-baselines library\cite{stable-baselines} with the policy network replaced by our custom policies described in section \ref{section:policies}. Before training, the hyperparameters were tuned using OpenTuner\cite{10.1145/2628071.2628092} with a custom script that can be found in our repository. The hardware used for all the training and experiments was a single machine running Ubuntu Linux 18.04 with access to 6 CPU cores (no GPU was used as the LP step makes the process CPU-bound).

\subsection{Experiments}
\begin{figure}
    \centering
    \resizebox{0.45\textwidth}{!}{\input{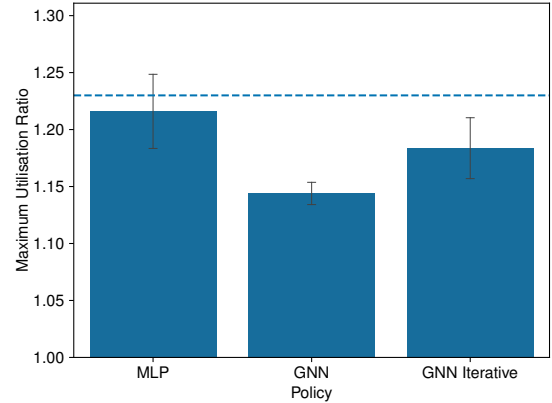}}
    \caption{Learning to route on a fixed graph. Bar heights are the mean ratio between achieved max-link-utilisation and the optimal for the given DM. The dotted line is the utilisation achieved by shortest-path routing. Lower is better.}
    \label{fig:fixed_plot}
\end{figure}

Our first experiment was designed to show that the GNN method does not cause any extra overhead when compared to the MLP method for situations where both are applicable. We ran both on the Abilene graph from Topology Zoo on sequences of 60 DMs with a cycle length of 10 and memory length of 5. 7 sequences were used for training with another 3 for testing. The reason these specific values were chosen is that they were used for the main example in previous work\cite{valadarsky2017learning} and so provide the best comparison. The results can be seen in Figure~\ref{fig:fixed_plot}. We can see that all policies outperform shortest-path routing, and that the GNN policies have the best performance. This is likely because the GNN policies only allow communication between related nodes and edges, and are less likely to overfit.

\begin{figure}
    \centering
    \resizebox{0.45\textwidth}{!}{\input{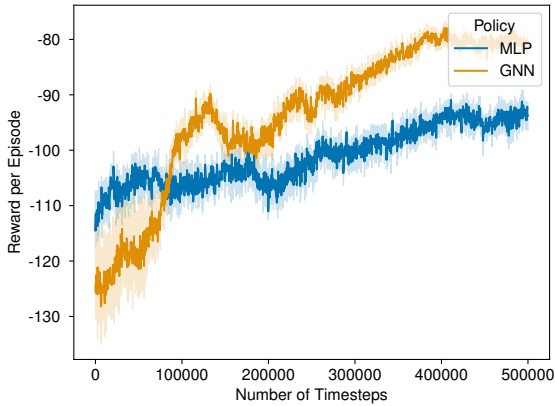}}
    \caption{Learning curves for the training of both MLP and GNN agents. The line is the mean of the total reward per episode throughout the training of an agent and the paler block is a confidence interval. The higher the reward, the better the agent has performed.}
    \label{fig:learning_curves}
\end{figure}

Our second aim was to show that there are no extra overheads in the learning process itself. To do so, we have generated Figure~\ref{fig:learning_curves} showing the learning curves for both agents. Both agents learnt at the same rate of roughly 70 frames per second, meaning the total time for 500000 training steps was 2 hours on a commodity PC. We can see that both policies do learn, but the GNN policy learns faster and finishes with better performance.

\begin{figure}
    \centering
    \resizebox{0.45\textwidth}{!}{\input{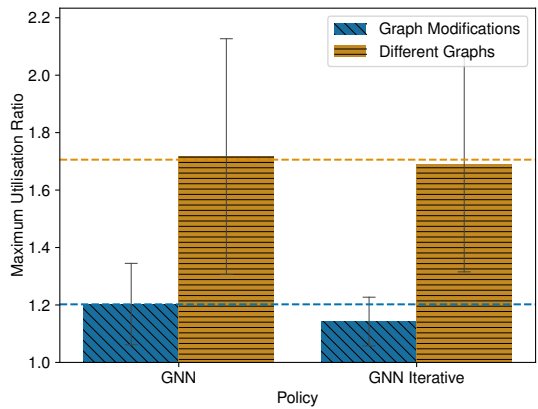}}
    \caption{Generalising to unseen graphs. Bar heights are the mean ratio between achieved max-link-utilisation and the optimal for the given DM. The dotted line is the utilisation achieved by shortest-path routing. Lower is better.}
    \label{fig:generalise_plot}
\end{figure}

Finally we aimed to show that the GNN policies can achieve their main goal: to enable generalisation of strategies to different network topologies. We ran a similar experiment to the first one, but this time just comparing our GNN and GNN-Iterative policies and training and testing on both a mixture of entirely different graphs from the dataset (between double and half the size of the Abilene graph), and the same graph with small modifications. The small modifications were the addition or deletion of one or two edges or nodes (chosen randomly). The results of this can be seen in Figure~\ref{fig:generalise_plot}. We can see that both policies generally perform at least as well as the baseline. However, as expected, the iterative policy has better performance. One oddity is the very different heights of the two bars. This is because for some graph structures, the approximations that softmin routing makes means it is much harder to get as close to the multipath routing optimum. The set of `different graphs' contained such examples, whereas changes to the Abilene graph did not cause this to occur.

\section{Discussion}
\label{section:discussion}
Our original aim was to explore whether GNNs, when combined with DRL, can provide benefits to work on systems, in particular our example of routing. Above, we have performed several experiments and given their results to show that these benefits are indeed real. The first of the experiments was designed to compare our work against the preceding seminal work on which much of it was based. In the Figure~\ref{fig:fixed_plot} we can indeed see that the GNN approach achieved better performance than the MLP approach and as they were trained for the same number of steps this means that the GNN does not lead to the agent being any less expressive. Furthermore, we can see in Figure~\ref{fig:learning_curves} the learning curves of both policy architectures. From this graph it is obvious that the GNN architecture reached a plateau first (although it did start off worse), which implies that it can learn more quickly to solve this problem than the MLP architecture, again supporting our argument that the more complex policy does not add a meaningful overhead. Finally, Figure~\ref{fig:generalise_plot} shows that the GNN architectures can generalise and are almost as performant when the topology changes. Also, as expected, the iterative version performs better as it was better designed for changing topologies.

We also believe that using GNNs for policy architectures where the problem involves graphs with changing topologies greatly simplifies the policy design procedure, making it easier for researchers and end users to build agents to optimise their systems. This is because the unit that the GNN operates on at the most basic level is a graph meaning that there is no extra work required to transform for input or output constraints. Clearly, an MLP architecture is far simpler still and can be used on such problems, but as demonstrated in this paper it has limitations and so is ineffective where generalisation is required. The only other option then is to build a custom policy by hand, for example based on LSTMs like that designed in Placeto\cite{addanki2019placeto}. However, this is far more time-consuming and prone to errors.

Of course, there is already a large body of research into GNNs and their properties. As such, many of the benefits described therein are instantly applicable here. It is well-known that GNNs allow for the careful specification of relations between data being learned on (implicitly in the graph), both being constrictive enough to allow only meaningful data flows but flexible enough to allow for complex structures (unlike CNNs for example)\cite{47094}. As networks are already graphs and the important relations are implicit in this structure, we get to take advantage of the induced biases for free, which we propose is why performance is not lost in comparison to the MLP even though MLPs are denser. Furthermore, we propose that GNNs being less dense, having only relevant relations, plays a role in why the learning was more efficient for the GNN-based policy. Also, due to the nature of the generalisability of GNNs, they are far more scalable than the MLPs here when it comes to learning as the parameter count for GNNs remains fixed with larger graphs but in the MLP case continues to grow.

The work here just scratches the surface of how powerful GNNs can be as a policy architecture in this field. All of the benefits mentioned above that led to good results for the routing example are inherent on the architecture of the GNN and so will be applicable to other problems involving networks. GNNs are still a very active research area meaning that new types of GNNs with different focusses are continually being created and honed, improving attributes such as their scalability and flexibility, removing roadblocks to further use cases that may have existed previously.

\subsection{Further Work}
Given the power of GNNs and that they have been shown here to provide generalisation benefits and more, it seems sensible that this line of research be continued. One useful approach would be to improve on work on routing in this paper. A core element of our approach focussed on designing a method to reduce the size of the action space. An important part of this was in designing a mapping from edge weights to a routing strategy. We believe that an exploration of different techniques in mapping edge weights, or indeed any intermediate structure, to a routing strategy could provide interesting results.

While the size of the action space was an issue we faced and tried to fix with a custom method, it is a major problem faced by DRL techniques more generally. Many applications have very large action spaces which cause issues for successful exploration and learning. An area of research that may provide solutions to this problem is that of extracting hierarchical structure from action spaces with pioneering works such as `A Hierarchical Mode for Device Placement'\cite{mirhoseini2018a} and `Learning Index Selection with Structured Action Spaces'\cite{welborn2019learning}. We believe that using similar ideas to these works could solve some of the problems we faced, and that further work in this area is critical for developing GNN DRL techniques that require less custom configuration and work.

Furthermore, how demand information is input to the network could be greatly improved and hierarchical DRL could be explored here. There are also many different ways in which the learning itself could be modified. These include using different learning algorithms to PPO. Another aspect that could be looked into are modifications to the reward function to better aid exploration.

The routing can also be expanded to examine different utility function and natural next step would be implementing these strategies in real-world SDN systems so that they could be tested and used on real-world networks.

Outside the area of data-driven routing there are likely to be many other useful applications to explore. Other aspects of network control have been investigated with ML and particularly DRL techniques such as resource scheduling and rate controls which may benefit greatly from the application of GNN-based techniques.

\section{Related Work}

There have been many previous projects that have sought to improve link congestion in networks with different routing strategies but only more recently have we started to see so many ML-based approaches.

\subsection{Oblivious Approaches}
\label{section:oblivious}

The first set of approaches are so-called oblivious\cite{bansal2008} approaches which do not take into account the traffic and instead try to create a routing strategy that provides good results for any demand matrix.

Early examples of such approaches are that of R\"acke\cite{1181881} who showed that it is possible to have an optimal oblivious routing, and Azar et al.\cite{10.1145/780542.780599} who proved that this can be calculated in polynomial time. Since then, small, incremental improvements have been made, improving calculation time or allowing for different utility functions to be specified\cite{kodialam2008advances}.

However, it is still possible to do better than oblivious routing if one has some knowledge of the traffic. One seminal work in this area is SMORE\cite{10.1145/3232755.3232781} which combines oblivious routing and adaptive sending rates to improve reliable network performance under all kinds of operating conditions.

\subsection{Reinforcement Learning in Routing}

There have thus far been multiple approaches to network-related problems using machine learning and especially DRL techniques\cite{8701570}. There is a growing body of work covering areas from routing to deep packet inspection. Specifically, in the area of routing, RL research has been taking place for quite some time with an early seminal paper introducing the idea of Q-routing in 1993\cite{boyan1993packet} which attempts to use distributed on-device agents to route packets to minimise delay. Many pieces of work have followed a similar vein, gradually improving on Q-routing with ever more complex policies\cite{9149287}.

A separate issue is the traffic engineering problem where one can make global decisions for a large scale network (such as an autonomous system) on a longer timescale. In this area there is notably the work: `Learning to Route with Deep RL'\cite{valadarsky2017learning} which has been discussed in depth in this paper.

Similarly, another paper: `A Deep-Reinforcement Learning Approach for Software-Defined Networking Routing Optimisation'\cite{stampa2017adl} sought to use DRL to find good routes for a given traffic matrix.

A further work in this area is `A Multi-agent Reinforcement Learning Perspective on Distributed Traffic Engineering'\cite{9259413} which, like our work, is a data-driven approach to traffic engineering using DRL and seeks to reduce congestion. Where it differs is in the use of distributed agents to control traffic in different regions and only act on local data.

\subsection{GNNs in Routing}

Beyond the application of DRL to routing, there has more recently been interest in applying more specifically GNN-based methods to this domain. One such work in fact sought to improve QoS on a network using ML while being robust to topology change: `Network Routing Optimization Based on Machine Learning Using Graph Networks Robust against Topology Change'\cite{9016573}. Like this work, a GNN is used to be able to generalise in the case of topology change. However, the focus is on using data of the state of the network as it is currently, as opposed to predictions in advance.

Another work: `Unveiling the potential of Graph Neural Networks for network modelling and optimization in SDN'\cite{10.1145/3314148.3314357} similarly uses GNNs to generalise to different topologies, and aims to use traffic data to inform routing decisions. However, it differs in that it looks at the problem of reducing delay and jitter in the network.

\section{Conclusions}
In this work, we explored the benefits of using GNNs in combination with DRL when applied to systems. In particular we took the example of routing and showed that using GNNs indeed can allow us to generalise the solution of learning data-driven routing for a single graph to multiple topologies. We also showed that this benefit does not come with a downside of extra overheads. We also developed an open source RL environment which can be used for further experimentation on GNNs in routing.

We believe it is important that the combination of GNNs with DRL is explored further and taken up in future research, as it is applicable to many other problem domains where it can surely provide the same benefits and ease of generalisation as discussed in this work.

\section*{Acknowledgment}
The authors would like to thank Kai Fricke for his input on GDDR project. This research was partly funded by the Alan Turing Institute.

\bibliographystyle{IEEEtran}
\bibliography{refs}

\end{document}